\documentclass[acmsmall,10pt]{acmart}

\AtBeginDocument{%
  }

\usepackage{amsmath}
\usepackage{amsfonts}
\usepackage{amsthm}
\usepackage{color}
\usepackage{subfig}
\usepackage{url}
\usepackage{comment}
\usepackage{graphicx}
\usepackage{enumitem}

\usepackage{csquotes}
\usepackage{balance}
\usepackage[nolist,nohyperlinks]{acronym}
\usepackage{array}

\usepackage{multicol}
\usepackage{wrapfig}
\usepackage[normalem]{ulem}

\usepackage[ruled,linesnumbered]{algorithm2e}
\usepackage{bm}
\usepackage{xspace}
\newcommand{\ie}{\textit{i.e.}\xspace}
\newcommand{\eg}{\textit{e.g.}\xspace}

\newcommand{\og}{OGB$_{\text{cl}}$\xspace}
\newcommand{\ognew}{OGB\xspace}

\DeclareMathOperator{\radius}{radius}
\newcommand{\f}{\bm{f}}

\newcommand{\etal}{\emph{et al.}\xspace}

\usepackage{booktabs}

\DeclareRobustCommand{\bigO}{%
  \text{\usefont{OMS}{cmsy}{m}{n}O}%
}

\begin{document}

\title{An Online Gradient-Based Caching Policy with Logarithmic Complexity and Regret Guarantees}

\author{Damiano Carra}
\affiliation{%
  \institution{University of Verona}
  \country{Italy}
}

\author{Giovanni Neglia}
\affiliation{%
  \institution{Inria, Universit\'e C\^ote d'Azur}
  \country{France}
}

\renewcommand{\shortauthors}{Carra et al.}

\begin{abstract}
Commonly used caching policies, such as LRU (Least Recently Used) or LFU (Least Frequently Used), exhibit optimal performance only under specific traffic patterns. Even advanced machine learning-based methods, which detect patterns in historical request data, struggle when future requests deviate from past trends. Recently, a new class of policies has emerged that are robust to varying traffic patterns. These algorithms address an online optimization problem, enabling continuous adaptation to the context. They offer theoretical guarantees on the \emph{regret} metric, which measures the performance gap between the online policy and the optimal static cache allocation in hindsight. However, the high computational complexity of these solutions hinders their practical adoption.

In this study, we introduce a new variant of the gradient-based online caching policy that achieves groundbreaking logarithmic computational complexity relative to catalog size, while also providing regret guarantees. This advancement allows us to test the policy on large-scale, real-world traces featuring millions of requests and items—a significant achievement, as such scales have been beyond the reach of existing policies with regret guarantees. To the best of our knowledge, our experimental results demonstrate for the first time that the regret guarantees of gradient-based caching policies offer substantial benefits in practical scenarios.
\end{abstract}





\maketitle

\section{Introduction}

Caching is a fundamental building block employed in various architectures to enhance performance, from CPUs and disks to the Web. Numerous caching policies target specific contexts, exhibiting different computational complexities. These range from simple policies like Least Recently Used (LRU), First In First Out (FIFO)~\cite{yang2023fifo} or Least Frequently Used (LFU)~\cite{matani20211} with constant complexity to more sophisticated ones like Greedy Dual Size (GDS) \cite{cao1997cost} with logarithmic complexity, up to learned caches \cite{rodriguez2021learning} that necessitate a training phase for predicting the next request.

All caching policies, implicitly or explicitly, target a different traffic pattern. For instance, LRU and FIFO favor recency, assuming requests for the same item are temporally close, while LFU performs well with stationary request patterns \cite{romano2008quantitative}. Learned caches \cite{rodriguez2021learning}, which use Machine Learning tools to predict the next request and therefore improve the hit ratio, identify patterns in historical request sequences, and assume the same pattern will persist. In a dynamic context with continuously changing traffic patterns, none of these policies can consistently provide high performance, and it is indeed easy to design adversarial patterns that may hamper the performance of a specific traffic policy~\cite{paschos2019learning, bhattacharjee2020fundamental}.

Some recent caching policies \cite{paschos2019learning, bhattacharjee2020fundamental, paria2021lead, mhaisen2022online, mhaisen2022optimistic, si2023no} have been designed to be robust to any traffic pattern, making no assumptions about the arrival process. These policies are based on the \emph{online optimization} \cite{shalev2012online} framework. The key performance metric in this context is \emph{regret}, which is the performance gap (e.g., in terms of the hit ratio) between the online policy and the optimal static cache allocation in hindsight. Online caching policies aim for sub-linear regret with respect to the length of the time horizon, as this guarantees that their time-average performance is asymptotically at least as good as the optimal static allocation with hindsight. Policies with sub-linear regret are typically referred to as \emph{no-regret} policies.

\begin{figure*}
\begin{minipage}[t]{.45\linewidth}
  \vspace{0pt}  
    \centering
    \includegraphics[width=0.9\linewidth]{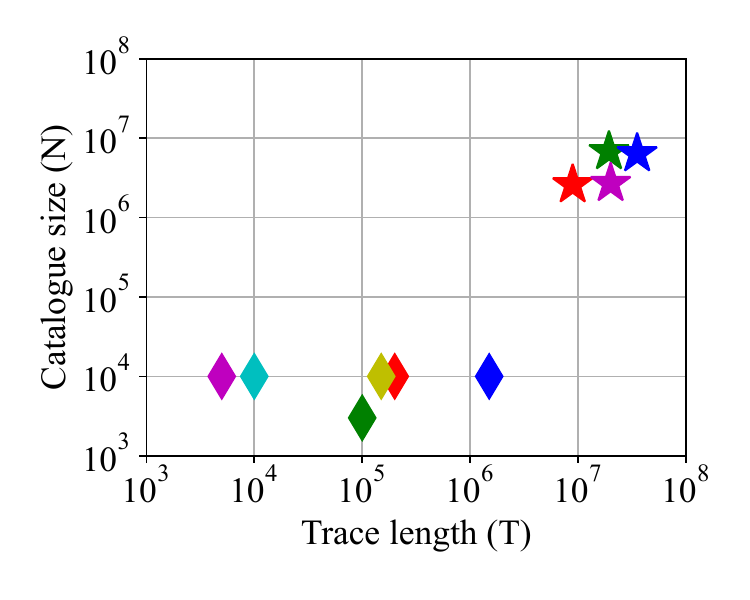}
    \vspace{-5mm}
    \captionof{figure}{Trace length ($T$) and catalog size ($N$) used in no-regret caching papers ($\blacklozenge$), and used commonly for evaluating caching policies ($\bigstar$).}
    \label{fig:trace_comparison}
\end{minipage}%
\hfill
\begin{minipage}[t]{.53\linewidth}
  \vspace{0pt}
    \centering
    \captionof{table}{Trace references.}
    \label{tab:trace_comparison}
    \vspace{-3mm}
    {\small
    \begin{tabular}{@{}llr|r|l@{}}
    \toprule
      ~ & {\bf Paper} & ~ & {\bf Year} & {\bf Label}  \\
    \hline
    $\color{red} \blacklozenge$ & Paschos\,\etal & \cite{paschos2019learning} & 2019 &  \texttt{no-regr$_1$} \\
    $\color{teal} \blacklozenge$ & Bhattacharjee\,\etal & \cite{bhattacharjee2020fundamental} & 2020 &  \texttt{no-regr$_2$} \\
    $\color{blue} \blacklozenge$ & Paria\,\etal & \cite{paria2021lead} & 2021 & \texttt{no-regr$_3$} \\
    $\color{magenta} \blacklozenge$ & Mhaisen\,\etal \,\,\,(a) & \cite{mhaisen2022online} & 2022 & \texttt{no-regr$_4$} \\
    $\color{cyan} \blacklozenge$ & Mhaisen\,\etal \,\,\,(b) & \cite{mhaisen2022optimistic} & 2022 &  \texttt{no-regr$_5$}\\
    $\color{yellow} \blacklozenge$ & Si Salem\,\etal & \cite{si2023no} & 2023 & \texttt{no-regr$_6$}\\
    \hline
    $\color{red} \bigstar$ & Kavalanekar\,\etal & \cite{kavalanekar2008characterization} & 2007 & \texttt{ms-ex} \\	
    $\color{teal} \bigstar$ & Lee\,\etal & \cite{Lee2017Understanding} & 2016 & \texttt{systor} \\	
    $\color{blue} \bigstar$ & Song\,\etal & \cite{song2020learning} & 2019 & \texttt{cdn} \\	
    $\color{magenta} \bigstar$ & Yang\,\etal & \cite{yang2020large} & 2020 & \texttt{twitter} \\	
    \bottomrule
    \end{tabular}
    }

\end{minipage}
\end{figure*}

\vspace{1mm}
\noindent
{\bf Limitation of the prior work.}
The major drawback of the no-regret policies is their computational complexity. Indeed, most of these policies have at least $\bigO(N)$ complexity per request, where $N$ is the catalog size~\cite{si2023no}. This high complexity has limited the adoption and  \emph{experimental validation}  of no-regret regret policies. Figure\,\ref{fig:trace_comparison} shows the trace length $T$ and the catalog size $N$ used in the literature proposing no-regret regret policies versus those in the more broader literature on caching policies (the corresponding references are listed in Table\,\ref{tab:trace_comparison}). No-regret policies have typically been evaluated using traces (labeled  \texttt{no-regr$_n$}) that are either generated synthetically, or sub-sampled from  much larger traces. Their catalogs and trace lengths are order of magnitude smaller than those considered in the rest of the literature.

Being able to provide theoretical guarantees may justify an increased computational complexity compared to basic policies such as LRU and FIFO. However, a $\bigO(N)$ complexity hinders the practical adoption of these policies in real-world scenarios. To the best of our knowledge only the Follow The Perturbed Leader (FTPL) policy \cite{mhaisen2022optimistic} can be, in some cases, implemented with $\bigO(\log N)$ complexity---see Sec.\,\ref{sub:motivation}. However, in practice, this policy is a noisy version of LFU, and tends to work well only when LFU does. In any case, FTPL has been evaluated, like other no-regret policies, on short traces with small catalogs---see Table\,\ref{tab:trace_comparison} and Fig.\,\ref{fig:trace_comparison}.

\vspace{1mm}
\noindent
{\bf Contributions.}
In this work, we introduce a new online caching policy, \ognew, based on the \emph{gradient descent} approach that has $\bigO(\log N)$ amortized computational complexity per request, and sub-linear regret guarantees in the \emph{integral caching} setting, i.e., when the cache stores entire items. This complexity matches that of commonly used caching policies, making it suitable for real-world contexts. Our solution relies on the clever joint design of two steps: (i) the selection of item storage probabilities, for which we modify the classic Online Gradient-Based (\og) policy originally proposed for \emph{fractional caching}—when the cache can store arbitrarily small fractions of every item—and (ii) a sampling process in which we select the actual set of items stored in the cache, minimizing the number of new items retrieved upon each update.

Our solution also works when the cache can be updated only after a \emph{batch} of requests are served. This case could be of interest in high-demand settings, or to amortize the computational cost of the caching policy and/or to reduce the load on the authoritative content server. The design maintains logarithmic complexity independently of the batch size.

By removing the sampling process, our policy also works in the fractional caching setting. It maintains regret guarantees, but its amortized complexity becomes $\bigO(N/B)$ per request, where $B$ is the batch size, matching state-of-the-art no-regret policies in the fractional setting.

The low complexity of \ognew allows us to evaluate it on traces with millions of requests and items---an accomplishment previous works could not achieve. We explore various scenarios where \ognew outperforms traditional caching policies, demonstrating that gradient-based schemes adapt effectively to changes in traffic patterns.

In summary, our contributions are as follows:
\begin{itemize}[topsep=0pt, noitemsep, leftmargin=*]
    \item We propose \ognew, the first integral online gradient-based caching policy that enjoys both sub-linear regret guarantees and $\bigO(\log N)$ amortized computational complexity per request, and can then handle typical catalogs that characterize real-world applications.
    \item We extend these results to  fractional caching  showing that our policy enjoys regret guarantees and  $\mathcal O(N/B)$ amortized complexity when operating over batches of $B$ requests, allowing to mitigate the linear complexity intrinsically associated with fractional caching.
    \item Through experiments on different public request traces with millions of items and tens of millions of requests, we show for the first time that the regret guarantees of online caching policies bring significant benefits in scenarios of practical interest.
\end{itemize}

\vspace{1mm}
\noindent
{\bf Roadmap.} 
The remainder of the paper is organized as follows. In Sec.\,\ref{sec:background}, we provide background information on gradient-based caching policies and discuss the motivation behind the work. In Sec.\,\ref{sec:overview}, we offer a high-level description of the elements composing our solution, and we prove its regret guarantees. Sections\,\ref{sec:projection} and\,\ref{sec:selection} provide detailed descriptions of the building blocks of our scheme. In Sec.\,\ref{sec:experiments}, we apply our solution to a set of publicly available traces. Finally, in Sec.\,\ref{sec:related} and Sec.\,\ref{sec:conclusion}, we discuss related work and conclude the paper.

\section{Background and motivation}
\label{sec:background}

\subsection{Online Gradient Based caching policy}
No-regret caching policies make no assumptions about the arrival pattern and adapt dynamically to the received requests. They aim to solve an \emph{online convex optimization} problem \cite{bhattacharjee2020fundamental}. We consider a catalog of $N$ items, $\mathcal{N} = {1, 2, \dots, N}$, all of equal size, and a cache with a capacity of $C < N$ items. We let the current time $t$ coincide with the number of requests received so far. Each request is represented as a one-hot vector $\bm{r}_t$, with $r_{t,j} = 1$ for the requested item $j \in \mathcal{N}$, while all other components are set to zero. 

We consider a batch operation, where the cache is allowed to update its state every $B \ge 1$ requests, as in~\cite{faticanti2024optimistic,si2023no}. The batched operation may be motivated by the fact that requests indeed arrive in batches, or by the need to amortize the computational cost of the caching policy and/or to reduce the load on the authoritative content server. In the first case, the ordering of $B$ consecutive requests at times $nB+1, nB+2, \dots, (n+1)B$ for $n \in \mathbb{N}$ is arbitrary.

We start describing the \emph{fractional setting}, considered for example in~\cite{bansal08, wang15, ji15, paschos2019learning}, where the cache can store arbitrarily small fractions of each item. The fractional setting is useful for modeling caching systems that work with item chunks much smaller than the item size, as is the case in video caches..
The cache fractional state at time $t$ is described by the vector $\bm{f}_t$, which belongs to the feasible state space $\mathcal F =\{\bm{f} \in [0,1]^N : \sum_{i=1}^N f_i=C\}$.
Each component $f_{t,i}$ indicates the fraction of item $i$ stored in the cache and the fractions sum to the cache capacity $C$. 

For the $t$-th request, the cache receives a reward $\phi_t(\bm{f}_t) \triangleq \sum_{i=1}^{N} w_{t,i} r_{t,i} f_{t,i}$, where $w_{t,i}$ is a weight that may be correspond to the cost of retrieving the whole item from the origin server. Upon a request for item $i$ ($r_{t,i}=1$), $w_{t,i} f_{t,i}$ can be interpreted as the cost reduction due to the fact that the fraction $f_{t,i}$ of time $i$ is stored locally. For simplicity, in this paper we consider $w_{t,i} = 1$ $\forall t, i$, which corresponds to consider the (fractional) hits as reward, but our results can be easily be extended to the general setting. The performance of an online caching policy $\mathcal{A}$ is characterized by the static \emph{regret} metric, defined as:
\begin{equation}
\label{eqn:regret}
R_T\left(\mathcal{A}\right) \triangleq \sup_{\bm{r}_0, \bm{r}_1, \ldots, \bm{r}_{T-1}} \left\{ \sum_{t = 0}^{T-1} \phi_t(\bm{x}^*) - \mathbb{E}\left[\sum_{t = 0}^{T-1} \phi_t(\bm{f}_t)\right] \right\}
\end{equation}
where $\bm{x}^* = {\arg\max}_{\bm{x} \in \mathcal{F}} \sum_{t = 1}^T \phi_t(\bm{x})$ is the best-in-hindsight static cache allocation knowing all  future requests,\footnote{
    Note that one can always select $\bm{x}^*$ so that only full items are stored, i.e., $x^*_i \in \{0,1\}$ for all $i \in \{1, \dots, N\}$.
}
and the expectation is over possible random choices of the caching policy. Such an allocation $\bm{x}^*$ is usually referred to as OPT. The regret measures the accumulated reward difference between the baseline $\bm{x}^*$ and the online decisions $\{\bm{f}_t\}_{t=0,\dots, T-1}$ by algorithm $\mathcal{A}$. An algorithm is said to achieve sub-linear regret (or have no-regret), if the regret can be bounded by a sub-linear non-negative function of the time-horizon $T$. In this case, the ratio $R_T/T$ is upper-bounded by a vanishing function as $T$  diverges, meaning the time-average performance of the online policy is asymptotically at least as good as the performance of the best static caching allocation with hindsight.\footnote{
Note that the regret may also be negative as it is the case in Fig.~\ref{fig:recent_traces_res}, right.
} 

The first no-regret caching policy was proposed in~\cite{paschos2019learning} for the fractional setting and later extended to batched operation ($B>1$) and the integral setting in~\cite{si2023no}. We denote this "classic" policy by \og and introduce it directly in the batched setting. The cache is updated every $B$ requests at time instants $\mathcal{T}_B \triangleq {n B : n \in \mathbb{N}}$ as follows:
\begin{equation}
\bm{f}_{t} = \begin{cases}
    \Pi_\mathcal{F} \left( \bm{f}_{t-B} + \eta \sum_{\tau=t-B}^{t-1}\nabla  \phi_{\tau}(\bm{f}_{t-B}) \right), & \text{if } t \in \mathcal T_B,\\
    \bm{f}_{t-1}, & \text{o.w.},
    \end{cases}
\label{eqn:ogb_update}    
\end{equation}
where $\Pi_\mathcal{F}(\cdot)$ is the Euclidean projection onto the feasible state space $\mathcal{F}$,  
$\sum_{\tau=t-B}^{t-1}\phi_{\tau}(\bm{f}_{t-B})$ is the total reward accumulated by the cache for the previous $B$ requests, and $\eta \in \mathbb{R}^+$ is the learning rate. The learning rate is a parameter of the scheme, and selecting $\eta = \sqrt{\frac{C}{B T}\left(1-\frac{C}{N}\right)}$ guarantees that $R_T\!\left(\textrm{OGB}_{\textrm{cl}}\right) \le \sqrt{BT C \left(1- 
\frac{C}{N}\right)}$ \cite[Corollary~4.5]{si2023no}.

In the \emph{integral setting}, the cache is forced to store each item in its entirety. The cache state at time $t$ can then be characterized by a binary vector $\bm{x}_t \in \{0,1\}^N$. A hard cache capacity constraint imposes $\sum_{i=1}^N x_{t,i} = C$, but a soft cache capacity constraint has also  been considered in the literature, e.g.,~in \cite{fofack12,berger14,neglia18ton,dehghan19,carra20}, requiring that the constraint is satisfied only in expectation over some random choices of the caching policy, i.e. $\mathbb{E}\left[\sum_{i=1}^N x_{t,i} \right]=C$. Besides satisfying the cache size constraint, one would also like to minimize the changes between the current and the previous cache states. In LRU, for instance, at most one item is added and another item is removed, which has a limited impact on the amount of data exchanged between the cache and the origin server. 

The concept of regret can be immediately extended to the integer setting by replacing $\bm{f}_t$ in~\eqref{eqn:regret} by $\bm{x}_t$.
Interestingly, a no-regret fractional caching policy can be transformed in a no-regret integral caching policy by augmenting it with an item \emph{sampling} procedure that guarantees $\mathbb{E}\left[\bm{x}_t\right] = \bm{f}_t$~\cite[Sec.~6]{si2023no}. This sampling procedure is also often referred to as a \emph{rounding scheme}. We discuss different rounding schemes in Sec.\,\ref{sec:selection}. 

\vspace{1mm}
\noindent
{\bf Projection Time Complexity.} 
The most expensive operation in \og is the projection on the capped simplex $\mathcal F$ \eqref{eqn:ogb_update}~ \cite{wang2015projection}, required both in the fractional and the integral setting. 
After $B$ requests are processed, the vector $\bm{y}_t = \bm{f}_{t-B} + \eta   \sum_{\tau=t-B}^{t-1}\nabla \phi_{t-B}(\bm{f}_{t-B})$ does not belong to the feasible set $\mathcal F$ (some components of $\bm{f}_{t-B}$ have been increased) and needs to be projected back to $\mathcal F$. The projection corresponds to solving the following problem: 
\begin{equation}
\begin{aligned}
\min_{\bm{f} \in \mathbb{R}^N} \quad & \frac{1}{2}\lVert \bm{f} - \bm{y}_t\rVert^2 \quad \quad \textrm{s.t.} \quad 0 \le f_i \le 1 \forall i \in \{1, \dots, N\}, \quad \sum_{i=1}^{N} f_i = C.
\end{aligned}
\label{eqn:min_proj}
\end{equation}
For the general case in which more than one component of $\bm{y}_t$ differs from the current $\bm{f}_{t-B}$, the best known Euclidean projection algorithm has $\bigO(N^2)$ complexity \cite{wang2015projection}. 
If a single component changes with respect to the current $\bm{f}_t$, Paschos\,\etal \cite{paschos2019learning} show that the complexity can be reduced to $\bigO(N\log N)$.
The best known per-request amortized cost is then $\bigO(N^2/B)$ for $B>1$ and $\bigO(N\log N)$ for $B=1$. 
We observe that no-regret algorithms based on online mirror descent (OMD), rather than on classic gradient descent, rely on a different projection (Bregman projection) and may achieve $\bigO(N/B)$ per-request amortized cost~\cite{si2023no}. \og is shown to outperform OMD in most settings of practical interest~\cite{si2023no}.

\vspace{1mm}
\noindent
{\bf Sampling Time Complexity.} 
In the integral setting, item sampling incurs additional costs. 
A naïve solution to update the cache is to consider each item independently and select it with probability $f_i$. This approach satisfies the soft cache constraint but not the hard one. Moreover, it can potentially lead to changing a large number of items in the cache.
To select exactly $C$ items, it is possible to adopt the \emph{systematic sampling} approach devised by Madow \emph{et al.} \cite{hartley1966systematic}. While this represents an elegant solution to a complex issue, it still requires $\bigO(N)$ operations and does not guarantee minimizing the changes in the cache, although experimental evidence shows that the number of replaced items is indeed small~\cite{si2023no}.
By formulating the problem as an optimal transport problem, we can guarantee the selection of exactly $C$ items and minimize the number of replacements. However, this approach comes with a high computational cost, namely $\bigO(N^3)$ complexity \cite{peyre2019computational}.
Overall, it is not possible to simultaneously achieve these three desired properties---selection of exactly $C$ items, minimization of replacements, and low computational complexity.

Both building blocks described above have an amortized cost of $\Omega(N/B)$, which is linear in the catalog size. In what follows, we demonstrate that, in the integral setting with soft cache capacity constraints, a slight modification of the \og policy in~\eqref{eqn:ogb_update}, combined with a smart joint design of the projection and sampling steps, results in a per-request cost of $\bigO(\log N)$ for any batch size $B \ge 1$, while still providing the same regret bound. In the fractional setting, our policy incurs a cost of $\bigO(N/B)$, still improving upon \og by a factor of $\log N$ when $B=1$ and by a factor of $N$ when $B>1$.

\subsection{Motivation}
\label{sub:motivation}
\begin{wrapfigure}{r}{0.42\linewidth}
    \includegraphics[width=\linewidth]{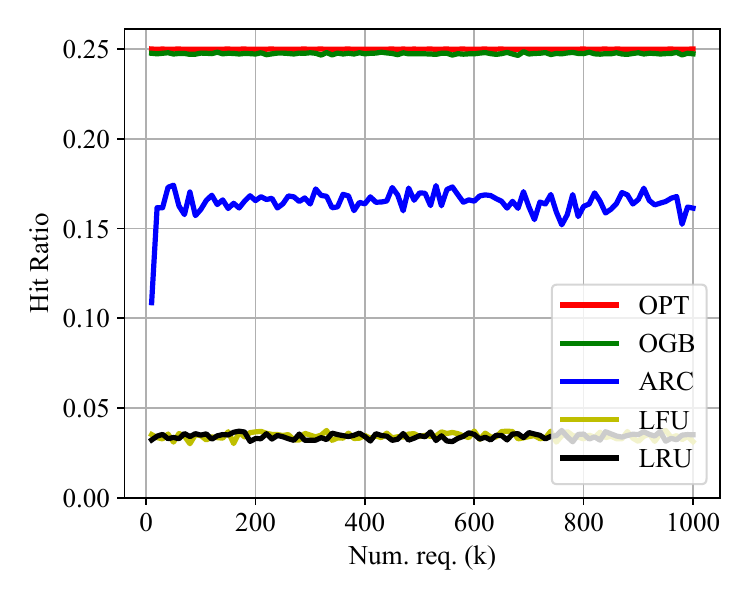}
    \caption{Adversarial trace. While recency or frequency based policies are not able to cope with adversarial patterns, online gradient based policies show a close-to-optimal performance.}
    \label{fig:adv_comp}
\end{wrapfigure}

\noindent
{\bf Adversarial trace.} 
We start considering an adversarial context in which the request pattern has been designed to limit the efficacy of the most common caching policies. In particular, we have a catalog with $N = 10^3$ items that are requested in a round-robin fashion. In each round, the request order is random. \ie, each round follows a different permutation of the item identifiers. The optimal static policy (OPT) arbitrarily selects $C$ items and leave them in the cache. In Fig.\,\ref{fig:adv_comp} we show the case with a cache size $C=250$, \ie, 25\% of the catalog. Unless otherwise stated, we consider a batch size of 1, in which case \og and our variant \ognew experience exactly the same number of hits.

Caching policies that favor recency (LRU) or frequency (LFU) are not able to exploit the caching space efficiently since in every round they replace most of the cache content. Even the ARC policy \cite{megiddo2003arc}, which dynamically decides the importance to give to recency and frequency, is not able to obtain the optimal cache hit ratio. Online gradient based policies, on the other hand, are able to obtain a close-to-optimal hit ratio, with a bounded error that depends on the learning rate $\eta$ \cite{si2023no}.

This simple adversarial trace demonstrates the limitations of known policies, which work well for specific arrival patterns but not in the general case. The example shows that LRU and LFU 
policies indeed have linear regret (the difference in the cumulative number of hits between OPT and LRU or LFU grows linearly with the number of requests), which was theoretically proven in \cite{paschos2019learning}.

\begin{figure}[th]
    \centering
    \includegraphics[width=.33\linewidth]{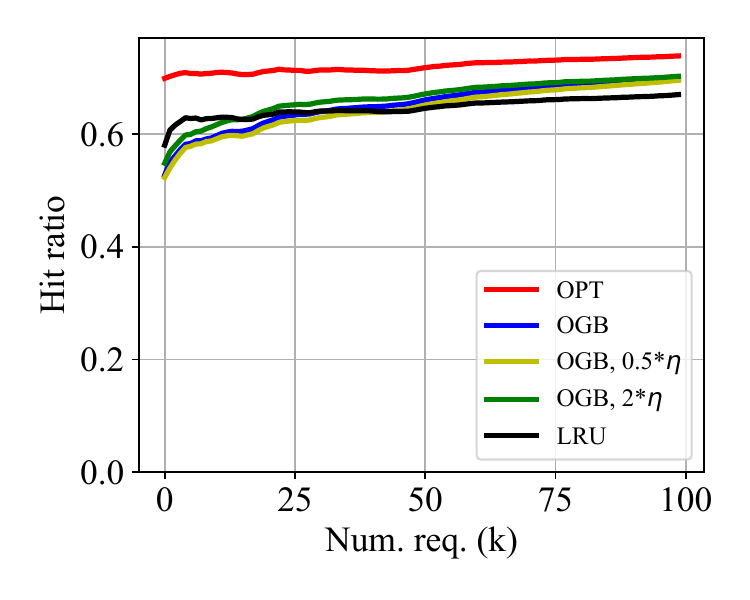}
    \includegraphics[width=.33\linewidth]{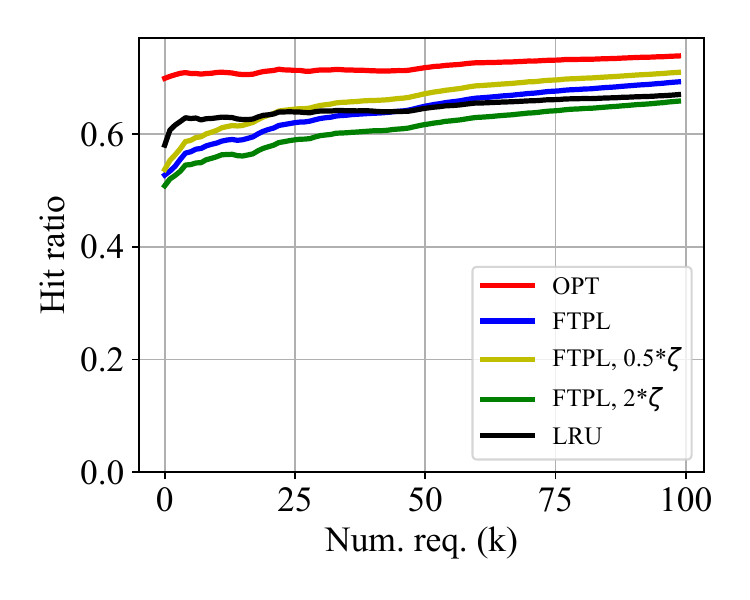}
    \caption{Real-world short traces: sensitivity of \og and FTPL. In \og, the parameter $\eta$ is the learning rate, while in FTPL the parameter $\zeta$ is the noise added to the LFU counters.}
    \label{fig:cdn_short_sensitivity}
\end{figure}

\vspace{1mm}
\noindent
{\bf Real-world (short) trace.} 
The adversarial setting may be considered unrealistic. Figure\,\ref{fig:cdn_short_sensitivity} (left), shows the results with a real-world trace. This trace contains $10^5$ requests for $10^4$ items (cache size $C = 500$ items) and it has been subsampled from the publicly available trace from \cite{song2020learning, cdnRepository} (trace with label \texttt{cdn} in Table\,\ref{tab:trace_comparison}), to generate a trace comparable, in terms of catalog size and number of requests, with the traces considered in previous works on no-regret policies (traces with labels \texttt{no-regr} in Table\,\ref{tab:trace_comparison}).
While LRU works reasonably well on this trace, it is outperformed by online gradient-based policies in the long term and, in any case, does not provide theoretical guarantees. 

\begin{figure}[t]
    \centering
    \includegraphics[width=.33\linewidth]{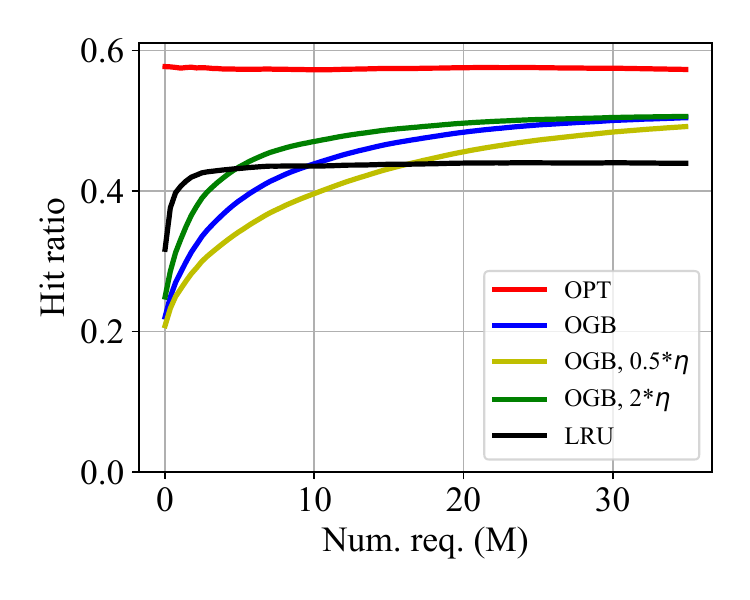}
    \includegraphics[width=.33\linewidth]{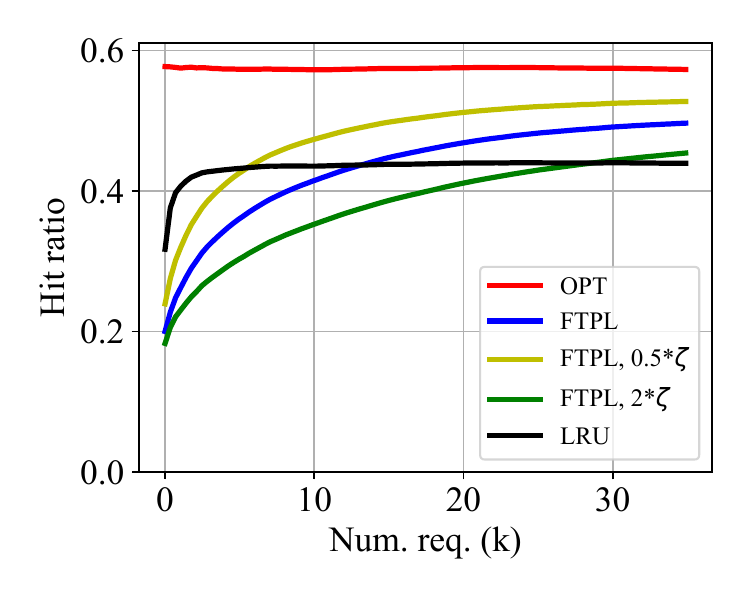}
    \caption{Real-world long traces: sensitivity of \og and FTPL. The initial noise added by FTPL heavily influences the performance.}
    \label{fig:cdn_long_sensitivity}
\end{figure}

Figure\,\ref{fig:cdn_short_sensitivity} (right) shows the performance of another no-regret policy, i.e., \emph{Follow The Perturbed Leader} (FTPL) \cite{bhattacharjee2020fundamental}. FTPL operates as LFU, but adds to each counter an independent Gaussian random variable with zero mean and standard deviation $\zeta$. By tuning $\zeta = \frac{1}{(4\pi \log N)^{1/4}}\sqrt{\frac{T}{C}}$, FTPL guarantees sublinear regret~\cite{bhattacharjee2020fundamental}. 
Figure\,\ref{fig:cdn_short_sensitivity} suggests that FTPL is much more sensitive to tuning $\zeta$ than online gradient based policies are to tuning $\eta$. This observation is confirmed by other experiments like those in Fig.\,\ref{fig:cdn_long_sensitivity}.

\vspace{1mm}
\noindent
{\bf Open issues.} 
The main issue with the \og policy is its $\bigO(N)$ computational complexity, which may limit it to a theoretical exercise with no practical application. To the best of our knowledge, no work shows that \og performs well with large catalogs and long traces that have heterogeneous patterns, like the ones used commonly in the caching literature (last 4 lines of Table\,\ref{tab:trace_comparison}).

Thanks to \ognew's reduced complexity of $\mathcal O(\log N)$, we can easily evaluate its performance on real-world traces with tens of millions of items and hundreds of millions of requests. For example, in Fig.\,\ref{fig:cdn_long_sensitivity}, we tested the trace \texttt{cdn} ($6.8 \cdot 10^6$ items and $3.5 \cdot 10^7$ requests) without the need to subsample it (as done in Fig.\,\ref{fig:cdn_short_sensitivity}).
Figure\,\ref{fig:cdn_long_sensitivity}, left, demonstrates that \ognew may provide better performance than classic policies like LRU.

To the best of our knowledge, the only existing no-regret policy with sublinear complexity is a variant of FTPL, where the Gaussian noise is generated only initially, instead of at each step (as originally proposed in~\cite{bhattacharjee2020fundamental}). This reduce FTPL's complexity from $\bigO(N\log N)$ (as the vector of counters need to be sorted at each request) to $\bigO(\log N)$~\cite{mhaisen2022optimistic}.
Nevertheless, the FTPL policy has some issues. FTPL is equivalent to LFU with some initial noise, which can be very large if the time horizon $T$ is large, causing it to suffer from the same limitations as LFU, such as poor adaptability to dynamic traffic patterns. Additionally, FTPL has a high sensibility to variations in its parameter $\zeta$, as already observed in Fig.~\,\ref{fig:cdn_short_sensitivity} and confirmed in Fig.~\,\ref{fig:cdn_long_sensitivity}, right. On the contrary, \ognew is robust to the setting of its parameter $\eta$ (Fig.~\,\ref{fig:cdn_long_sensitivity}, right). Moreover, our experiments in Sec.~\ref{sec:experiments} show that it is better suited than FTPL to track changes in request patterns.

\section{Solution overview}
\label{sec:overview}

\begin{figure}[h]
\begin{minipage}[t]{.45\linewidth}
  \vspace{0pt}
  {\footnotesize
  \begin{algorithm}[H]
    \SetKwInOut{Input}{input}
    \SetAlgoVlined

    \BlankLine
    $\bm{f}_0 \leftarrow \frac{C}{N}\bm{1}$\;
    ${\bm{x}}_0 \leftarrow \textsc{Bernoulli}(\bm{f}_0)$\;
    \For{ $t = 1, \ldots, T$}{
      $\bm{f}_{t} \leftarrow \textsc{UpdateProbabilities}(\bm{f}_{t-1},\bm{r}_{t-1})$\;
      \tcp{update according to \eqref{eqn:ogbnew_update}}
      
      \If{$t~\%~B == 0$ }{ 
        \tcp{batch processed}
        ${\bm{x}}_{t} \leftarrow \textsc{UpdateSample}(\bm{f}_{t},{\bm{x}}_{t-B})$\;
        \tcp{update the cache, $\mathbb{E}[\bm{x}_t]=\bm{f}_t$}
      }
    }
  \caption{\ognew scheme}
  \label{alg:overall_scheme}
  \end{algorithm}
  }
\end{minipage}
\hfill
\begin{minipage}[t]{.54\linewidth}
  \vspace{0pt}  
    \centering
    \includegraphics[width=1.0\linewidth]{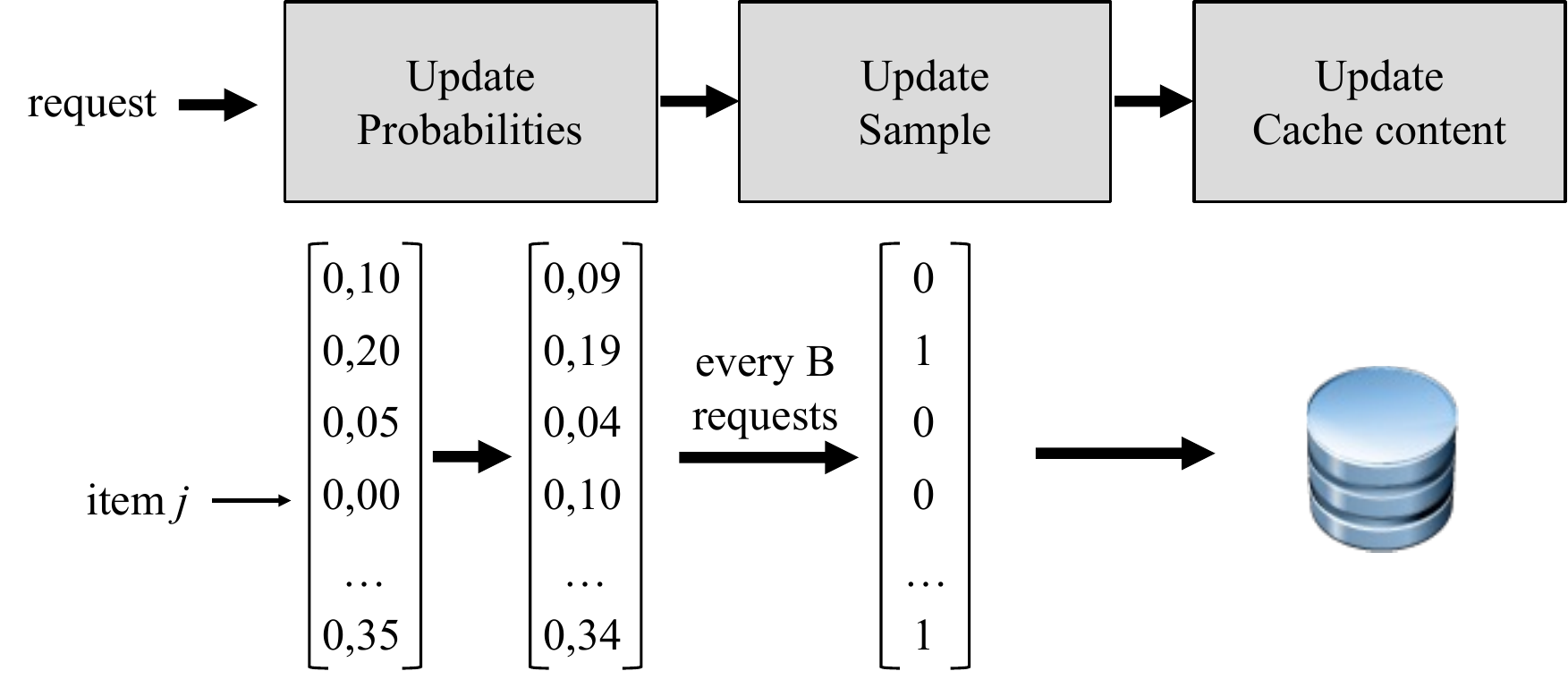}
    \vspace{-8mm}
    \captionof{figure}{High level view of the building blocks that compose the solution: the item selection passes through the computation of the caching probabilities.}
    \label{fig:overall_scheme}
\end{minipage}%
\end{figure}

Our scheme is presented in Algorithm~\ref{alg:overall_scheme} for the integral setting. The extension to the fractional setting is presented in Sec.~\ref{sec:fractional}. We denote our scheme as \ognew because it is almost identical to the \og policy in~\eqref{eqn:ogb_update}. Indeed, as \og, \ognew maintains a vector $\bm{f}_t$ indicating the probability with which each item is stored, and updates the integral cache status every $B$ requests according to $\bm{f}_t$, guaranteeing that $\mathbb{E}[\bm{x}_t]=\bm{f}_t$. The only difference is that, while \og updates the vector $\bm{f}_t$ once every $B$ requests, \ognew updates it after every request as follows:\footnote{Note that \og and \ognew coincide for $B=1$.} 
\begin{equation}
\bm{f}_{t} = 
    \Pi_\mathcal{F} \left( \bm{f}_{t-1} + \eta \nabla  \phi_{t-1}(\bm{f}_{t-1}) \right).
\label{eqn:ogbnew_update}    
\end{equation}

As we are going to show, \ognew achieves $\bigO(\log N)$ per-request  amortized cost for any $B\ge 1$ against \og's $\bigO(N \log N)$ and $\bigO(N^2/B)$ for $B=1$ and $B>1$, respectively.
This may appear surprisingly because \ognew performs  more projections and, as observed in Sec.~\ref{sec:background}, the projection is the most expensive step in \og. A first explanation is that projections of vectors perturbed in a single component are less expensive to compute ($\bigO(N \log N)$ vs $\bigO(N^2)$). However, this difference alone does not account for \ognew's $\bigO(\log N)$ complexity. The key lies in the joint design of the projection and the sampling steps which allows the new cache state $\bm{x}_{t}$ to be sampled at $t \in \mathcal T_B$ without needing to update the vectors of probabilities $\bm{f}_{t-B+1}, \dots, \bm{f}_{t}$---an operation which would require at least $\bigO(N)$ operations.

Before presenting \ognew's implementation in detail, we observe that \og and \ognew lead to different sequences $(\bm{f}_t)_{t \in \mathcal T_B}$. The first question, then, is whether \ognew enjoys the same regret guarantees of \og. The following proposition (proof in Appendix\,\ref{sec:proof_regret})
 confirms that this is the case.

\begin{theorem}
\label{thm:fractional_regret}
The policy \ognew in Algorithm\,\ref{alg:overall_scheme}  with $\eta = \sqrt{\frac{C\left(1- \frac{C}{N}\right)}{T B}}$ has regret upperbounded by 
$\sqrt{C\left(1- \frac{C}{N}\right) TB}$.
\end{theorem}

In the following, we summarize the challenges of each individual step, \textsc{UpdateProbabilities} and \textsc{UpdateSample}

\section{Updating the Probabilities}
\label{sec:projection}

When an item is requested, the caching strategy should increase the probability of that item being cached, while decreasing the other items' probabilities. These two operations are indeed accomplished in two different steps by the \ognew policy. First, we increment the component related to the requested item by a value equal to the step size $\eta$.\footnote{
Remember that, for simplicity, we consider we consider $w_{t,j}=1$ for every $t$ and $j$, but our results can be easily extended to the more general case.
} Then, the projection $\Pi_\mathcal{F}(\cdot)$ decreases all the components, such that their sum is equal to $C$. In order to design a low complexity algorithm, we need to understand how the projection actually operates.

\begin{figure}[th]
    \centering
    \includegraphics[width=.33\linewidth]{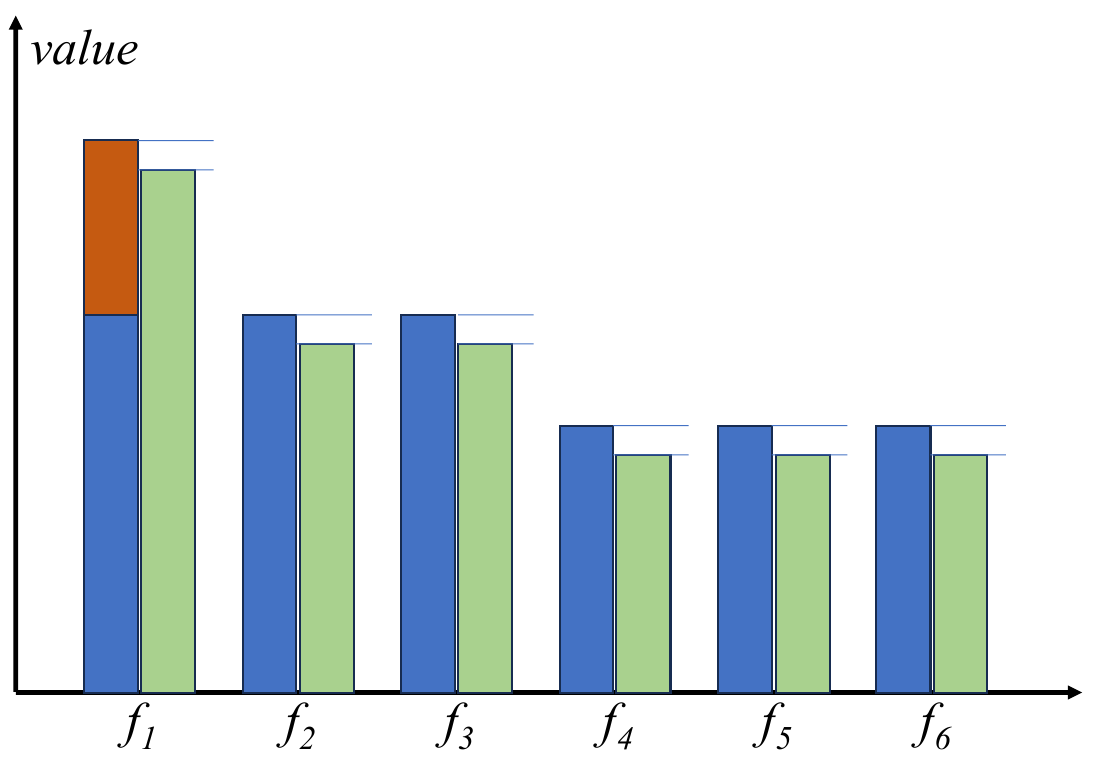}
    \caption{Example of a projection on a small vector. Starting from a feasible solution (blue bars), a new request for item 1 generates an excess (orange bar), which is taken evenly from all the components (green bars) to obtain a new feasible solution.}
    \label{fig:projection}
\end{figure}

From the previous vector $\bm{f}_{t-1}$, when item $j$ is requested, we obtain the following vector $\bm{y}_t$:
\begin{equation}
  y_{t,i} = \begin{cases}
     f_{t-1,i}  & \text{if } i \neq j,  \\
     f_{t-1,i} + \eta & \text{if } i = j.
  \end{cases}
  \nonumber
\end{equation}
Clearly we have $\sum_{i=1}^{N} y_{t,i} = C + \eta$ and the aim of the projection is to remove the excess $\eta$. Let $\mathcal{M}_p = \{i : y_{t,i} > 0\}$ be the set of indexes of the positive components of $\bm{y}_t$. In order to minimize the squared norm in Eq.\,(\ref{eqn:min_proj}), the excess should be \emph{uniformly taken from each positive component}. Let $\rho = \eta/|\mathcal{M}_p|$, then the projection $\bm{f}_{t}$ would be:
\begin{equation}
  f_{t,i} = \begin{cases}
     y_{t,i} - \rho   & \text{if } i \in \mathcal{M}_p,  \\
     0 & \text{otherwise}.
  \end{cases}
  \nonumber
\end{equation}
Note that the set $\mathcal{M}_p$ also includes the index of requested item $j$. As an example, Fig.,\ref{fig:projection} shows a small vector with six components. The initial state is represented by the blue bars. When a new request for item 1 arrives, we first add the orange contribution to component 1. We then uniformly decrease all the components (resulting in the green bars) to ensure the sum of all components remains constant.
To ensure that all components of the projection fall within the range [0,1], there are some corner cases that need to be considered.

The first case occurs when the component related to the requested item $j$ exceeds one, i.e., $f_{t-1,j} \le 1$, but $f_{t-1,j} + \eta - \rho > 1$. This case is easy to manage: the component $f_{t,j}$ is set to one, the excess becomes $\eta' = 1 - f_{t-1,j}$ and we remove $j$ from $\mathcal{M}_p$, \ie, $\mathcal{M}'_p = \mathcal{M}_p \setminus \{j\}$. Note that in the trivial case where $f_{t-1,j} = 1$, the $j$-th component of $\bm{y}_t$ becomes $y_{t,j} = 1 + \eta$, and the final projection is simply $\bm{f}_{t} = \bm{f}_{t-1}$.

The second case occurs when a component becomes negative, \ie, $y_{t,i} > 0$, but $y_{t,i} - \rho < 0$. This leads to changes in the set $\mathcal{M}_p$ and the excess to be redistributed. Specifically, $\eta' = \eta - \sum_{i : y_{t,i}<\rho}  y_{t,i}$, $\mathcal{M}'_p = \mathcal{M}_p  \setminus \{i : y_{t,i}<\rho\}$, and $\rho' = \eta'/|\mathcal{M'}_p|$. This adjustment may be needed multiple times, but, as we show in Sec.\,\ref{sec:proj_algo}, on average, a single item goes to 0 after each request.

While these corner cases require some attention, the main observation remains: the projection, in its essence, is a uniform redistribution of some excess. This aspect can be exploited in designing a low complexity algorithm for the update of the projection.

\subsection{Improving Projection: Main Idea}
\label{sec:main_idea}

Consider two items, $a$ and $b$, with $f_a > 0$ and $f_b > 0$.\footnote{
In the notation, we sometimes omit the time slot $t$ when it is clear from the context.
} Assume that they are not requested during two consecutive time slots and that their updated values do not become zero. After the first request, their values are decreased by the same quantity $\rho_1$. After the second request, their value are decreased by the same quantity $\rho_2$. In this case, we do not need to actually change $f_a$ and $f_b$, but we can maintain an external variable $\rho = \rho_1 + \rho_2$, knowing that the projections of the components $f_a$ and $f_b$ are indeed $f_a - \rho$ and $f_b - \rho$. We denote the ``unadjusted'' version of $\bm f$ by  $\bm{\tilde{f}}$, and keep track of the adjustment coefficient $\rho$ separately, so that %
\begin{equation}
  f_i = \begin{cases}
     \tilde{f}_i - \rho   & \text{if } \tilde{f}_i > 0,  \\
     0 & \text{otherwise}.
  \end{cases}
  \nonumber
\end{equation}

To manage the corner cases in the projection step, we need to detect when a component goes below zero. To this aim, we keep a copy of the (positive components of the) vector $\bm{\tilde{f}}$ sorted by value. When we update $\rho$ to a new value $\rho'$, we check which components are smaller than $\rho'$, remove them and adjust $\rho'$. When a request for item $j$ arrives, we update the corresponding value in $\bm{\tilde{f}}$ and in its sorted copy.

Managing the sorted copy of the coefficients has $\bigO(\log N)$ complexity. If we need the actual projection $\bm{f}$, we must compute each component from $\bm{\tilde{f}}$ and $\rho$. This implies that any algorithm updating the projection at each step cannot have a complexity less than $\bigO(N)$. However, in our case, since the projection is used to determine the probability of selecting items to cache, we do not need to update all components. We only need to identify which components fall below a given threshold, as explained in detail in Sec.\,\ref{sec:selection}.

\subsection{Improving Projection: The Algorithm}
\label{sec:proj_algo}

\begin{algorithm}
\begin{multicols}{2}
{\footnotesize
\SetKwInOut{Input}{input}
\SetAlgoVlined
\Input{ $\bm{\tilde{f}}$, current state}
\Input{ $\rho$, current adjustment}
\Input{ $\bm{z}$, ordered tree with positive coeffs. of $\bm{\tilde{f}}$}
\Input{ $j$, index of the requested item}
\Input{ $\eta$, \ognew step size}

\BlankLine
    \tcp{The requested item is already 1}
    \If{$\tilde{f}_j - \rho == 1$}{
        return\;
    }
    \tcp{The requested item was 0}
    \eIf{$\tilde{f}_j == 0$}{
        $\tilde{f}_j \leftarrow \rho + \eta$\;
        $z_j \leftarrow \rho + \eta$\;
        $\bm{z} \leftarrow \bm{z} \cup \{z_j\}$\;
    }{
        \tcp{Update with the \ognew step}
        $\tilde{f}_j \leftarrow \tilde{f}_j + \eta$\;
        $z_j \leftarrow z_j + \eta$\;
    }
    
    \tcp{Remove items with negative values}
    $\eta' = \eta$\;
    \Repeat{$\mathcal{B}$ is empty}{\label{alg:repeat}
        $\rho' = \eta'/|\bm{z}|$\;
        $\mathcal{B} \leftarrow \emptyset$\;
        $\mathcal{B} \leftarrow \{i : z_i - \rho - \rho' < 0\}$ \; 
        $\eta' \leftarrow \eta' - (z_i-\rho), \quad \forall i \in \mathcal{B}$\;
        $\bm{z} \leftarrow \bm{z} \setminus \{z_i : i \in \mathcal{B}\}$ \;
        $\tilde{f}_i \leftarrow 0, \quad \forall i \in \mathcal{B}$ \;
    } \label{alg:repeat_end}
    \tcp{Check the value of the requested item}
    \If{$(z_j \in \bm{z})$ and $(z_j - \rho - \rho'> 1)$}{ \label{alg:max}
        $\eta' = \eta - \left((z_j - \rho) - 1\right)$\;
        $\bm{z}, \bm{\tilde{f}} \leftarrow$ \textsc{RestoreRemoved}$()$\;
        $\bm{z} \leftarrow \bm{z} \setminus \{z_j\}$ \;
        $\tilde{f}_j  \leftarrow -1 $ \;
        GoTo line \ref{alg:repeat}
        \tcp*[l]{This can happen only once}
    } \label{alg:max_end}
    \tcp{Update $\rho$}
    $\rho \leftarrow  \rho + \rho'$\;
    \If{$z_j \notin \bm{z}$}{\label{alg:restore}
        $z_j \leftarrow 1 + \rho$\;
        $\bm{z} \leftarrow \bm{z} \cup \{z_j\}$ \;
        $\tilde{f}_j  \leftarrow 1 + \rho$ \;
    }\label{alg:restore_end}
    \Return $\rho, \bm{\tilde{f}}, \bm{z}$
	\caption{\textsc{UpdateProbabilities}}
	\label{alg:projection}
}
\end{multicols}
\end{algorithm}

The proposed Algorithm\,\ref{alg:projection} is based on the fact that the projection is a constrained quadratic program and the objective function is strictly convex with a unique solution characterized by the KKT conditions~\cite{nocedal1999numerical}. As done in \cite{wang2015projection} and \cite{paschos2019learning}, we look at the set of coefficients that are respectively greater than one, between zero and one, and zero, and we adjust them with the uniform redistribution of the excess. 

Differently from the above mentioned papers, following the ideas presented in Sec.\,\ref{sec:main_idea}, we maintain an ordered data structure $\bm{z}$, which contains the positive coefficients of the vector $\bm{\tilde{f}}$, and an external adjustment coefficient $\rho$. When a new request for item $j$ arrives, we first check if $\tilde{f}_j - \rho$ is already equal to one, in which case there is no update. Otherwise, we assign the step size $\eta$ to the component $j$. If $\tilde{f}_j$ was equal to zero, we adjust its value considering the current adjustment coefficient $\rho$.

The next block---lines \ref{alg:repeat}--\ref{alg:repeat_end}---handles the corner case in which the redistribution of the excess produces negative coefficients. Although this block may potentially be executed multiple times, we prove that, on average, a  single element will be set to 0 after each request, resulting in the block being  executed only once. 
Consider the cache after the first $t$ requests. At any time, there can be at most $N-C$ components of the vector $\bm{f}_t$ equal to 0. Moreover, at each request, at most one component can turn positive (because the corresponding item has been requested). It follows that the total number of components set to 0 over the $t$ requests is upper-bounded by $N-C+t$. Therefore, on average, $1 + \frac{N-C}{t}$ new components are set to 0 at each request, and the average time complexity of the cycle at lines \ref{alg:repeat}--\ref{alg:repeat_end} is $\mathcal{O}(1)$. In practice, our experiments show that the loop is executed at most twice per request, as also observed in~\cite[IV.B]{paschos2019learning} for the corresponding operations in their projection algorithm.

Lines \ref{alg:max}--\ref{alg:max_end} consider the corner case in which a coefficient is greater than one. There can be at most one such coefficient, corresponding to the requested item, therefore this block is executed at most once. If executed, we first derive the new excess to be distributed, and we restore the components removed in lines \ref{alg:repeat}--\ref{alg:repeat_end}. We take out the component of to the requested item (it will be restored at the end, at lines \ref{alg:restore}--\ref{alg:restore_end}), and we repeat lines \ref{alg:repeat}--\ref{alg:repeat_end} with the updated excess $\eta'$.

Overall, the time-complexity of projection update is then determined by the need to keep the data structure $\bm{z}$ ordered, which is $\mathcal O(\log N)$.

\section{Updating the Sample}
\label{sec:selection}
After the update of the caching probabilities, the cache may change the set of stored items. In the literature, this is often referred to as a \emph{rounding scheme} \cite{paria2021lead, si2023no}, since the fractional state $\bm{f}_t$ with continuous components between $0$ and $1$ is mapped to a caching vector $\bm{x}_t$ whose components are either zero (the item is not cached) or one (the item is cached).

The selection can be seen as a \emph{sampling process}, in which the probability that an item $i$ belongs to the sample is proportional to its component $f_i$. Considering the sampling literature, this is known as \emph{probability proportional to size} (PPS) sampling \cite{ohlsson1998sequential}. 

\vspace{1mm}
\noindent
{\bf Problem definition.} 
Since $\bm{f}$ varies as new requests arrive, we are interested in finding a scheme that, starting from the previous sample, and considering the updated $\bm{f}$, draws a new sample that has a large overlap with the previous one. Sample coordination refers to the process of adjusting the overlap between successive samples \cite{tille2020sampling}. With positive coordination the number of common items between two consecutive samples is maximized, while with negative coordination this number is minimized. We aim at finding a positive coordination design that is computationally efficient. 

\vspace{1mm}
\noindent
{\bf Current solutions.} 
PPS sampling schemes that provide an exact number of samples---such as systematic sampling used in \cite{paria2021lead, si2023no}---have $\bigO(N)$ complexity. To the best of our knowledge, there is no extension of systematic sampling to successive sampling that would allow for a reduction in the computational cost of consecutive sampling. The only option is to re-apply the scheme from scratch, without guarantees on the level of coordination across consecutive samplings.

Any scheme that aims at maximizing the overlap of successive samples with an exact sample size, requires the computation of joint inclusion probabilities over subsequent samples, which in turn involves an update on every pair of items \cite{tille2020sampling}, so it has at least $\bigO(N^2)$ complexity.

\vspace{1mm}
\noindent
{\bf Our approach.}
The conclusion is that under hard cache capacity constraints the sampling scheme complexity is $\bigO(N)$ and $\bigO(N^2)$ if we want to maximize the overlap between consecutive cache allocations.
We show that under a soft cache constraint, which allows the number of item to satisfy the contraint in expectation, we can design a low-complexity scheme with a tunable sample coordination.

\subsection{Coordination of Samples}
\noindent
{\bf First sample.} 
The components of $\bm{f}$ satisfy the following constraints: $0 \le f_i \le 1, \forall i \in \mathcal{N}$, and $\sum_{i=1}^{N} f_i = C$. Therefore, we can apply \emph{Poisson sampling} \cite{ohlsson1995coordination}, \ie, we decide to include an item in the sample independently from the other items. To this aim, we associate a random number $p_i$, drawn from a uniform distribution between zero and one, to each item $i$, and we include the item $i$ in the sample if and only if $p_i \le f_i$. The resulting sample has a random sample size with expectation $C$, the cache size. With sufficiently large $C$ the variability around the expectation is limited. In particular, the variability is the largest when each item has the same probability to be stored in the cache, i.e., $f_i = C/N$ for each $i \in \mathcal N$. Considering this setting, the coefficient of variation (the ratio between the standard deviation and the expected value) for the number of items sampled can be easily upper-bounded by $1/\sqrt{C}$. It is then smaller than 1\% as far as $C\ge 10000$. Moreover, under a Gaussian approximation, the probability that the number of items sampled exceeds the average cache size $C$ by $\epsilon C$ is upper-bounded by $\Psi(\epsilon \sqrt{C})$, where $\Psi(\cdot)$ is the complementary cumulative distribution function of a standard normal variable. For example, we can conclude that, in the same  setting, the probability that the instantaneous cache occupancy exceeds $C$ by more than 3\% is smaller than 0.13\%. Theorem 1 in \cite{dehghan19} presents a similar result based on the Chernoff bound.

\vspace{1mm}
\noindent
{\bf Subsequent samples.} 
Brewer\,\etal \cite{brewer1972selecting} showed that Poisson sampling guarantees positive coordination if the random value $p_i$ is \emph{permanently} associated to item $i$. The sample selection rule then becomes: the item is included as long as $p_i \le f_{t,i}$. 
Note that the values $\{p_i\}$ may periodically be randomly redrawn to reduce potential over/under-representation of some items.

\vspace{1mm}
\noindent
{\bf Putting the pieces together.} 
In Sec.,\ref{sec:projection}, we proposed a scheme where, instead of computing $f_i$ at every request, we update the global adjustment $\rho$, along with $\tilde{f}_i$ if item $i$ has been requested, knowing that $f_i = \tilde{f}_i - \rho$. After $B$ updates, we need to determine which items should be included in or evicted from the cache. We distinguish three groups of items.

The first group includes items that have been requested in the last $B$ requests. The value of $f_i$ for these items changed.
If the item is not currently cached,  we check whether $f_i \ge p_i$, i.e., $\tilde{f}_i - \rho \ge p_i$. If that is the case, we add it to the cache.

The second group contains the items that have not been requested and were not in the cache: since $f_i$ decreased, the items continue to stay out of the sample, \ie, we do not need to manage all the items not cached and not requested. 

The third group includes all the cached items (excluding the requested ones). We need to check if, for some items,  $f_i$ became smaller than $p_i$, \ie, $\tilde{f}_i - \rho < p_i.$ For these items, both $\tilde{f}_i$ and $p_i$ remained unchanged, only $\rho$ has been updated. The key observation is that, after $B$ updates, for all the cached 
\begin{wrapfigure}{r}{0.5\linewidth}
\begin{minipage}[t]{1.0\linewidth}
  \vspace{0pt}
  {\footnotesize
  \begin{algorithm}[H]
    \SetKwInOut{Input}{input}
    \SetAlgoVlined
    \Input{ $\bm{\tilde{f}}$, current fractional state}
    \Input{ $\bm{x}$, current cache state}
    \Input{ $\rho$, current adjustment}
    \Input{ $\bm{p}$, permanent random numbers}
    \Input{ $\bm{d}$, ordered tree with differences $(\tilde{f}_i-p_i)$}
    \Input{ $J$, set of indexes of the requested items}

    \BlankLine
    \tcp{Manage the requested items}
    \For{$j \in J$}{
        \eIf{$j$ \textnormal{index in} $\bm{d}$}{ \label{alg_insert}
            $d_j \leftarrow (\tilde{f}_j - p_j)$\;
        }{\If{$ \tilde{f}_j -  \rho \ge p_j $}{ 
              $d_j \leftarrow (\tilde{f}_j - p_j)$ \;
              $\bm{d} \leftarrow \bm{d} \cup \{d_j\}$\;
              \tcp{Add the item to the cache}
              $x_j = 1$\;
         }
       } \label{alg_insert_end}
    }
    \tcp{Eviction}
    $x_i = 0, \quad \forall i: d_i < \rho$\; \label{alg_evict}
    $\bm{d} \leftarrow \bm{d} \setminus \{d_i : d_i < \rho$\} \; \label{alg_evict_end}
	\caption{\textsc{UpdateSample}}
	\label{alg:selection}
  \end{algorithm}
  }
\end{minipage}
\end{wrapfigure}
items (except the ones that have been requested), the difference $d_i = \tilde{f}_i - p_i$ remains constant. We can then maintain the values $d_i$ in an ordered data structure and evict the items for which $d_i < \rho$ as $\rho$ changes.

\subsection{Item selection: The algorithm}

Algorithm\,\ref{alg:selection} summarizes the sampling operation to be executed every $B$ requests.
For each requested item $i$, if the item is already in the cache, we update the difference $d_i$. If it is not, we potentially insert it into the cache (lines \ref{alg_insert}-\ref{alg_insert_end}). Overall, we can update or insert at most $B$ values in the data structure.

Then, we evict all items whose difference $d_i$ is smaller than the current value of $\rho$. We can prove that, on average, $B$ elements need to be evicted (similarly to the proof that on average one component of $\bm f$ is set to 0 in Sec.~\ref{sec:proj_algo}). The ordered data structure $\bm{d}$ guarantees $\bigO(\log N)$ complexity for each eviction. Overall, we achieve the target $\bigO(\log N)$ per-request amortized complexity for the sampling procedure.

\subsection{Fractional Caching}
\label{sec:fractional}

In the fractional setting, item selection is straightforward: since $\sum_{i=1}^{N} f_{t,i} = C$, we cache a fraction $f_{t,i}$ of each item $i$ for which $f_{t,i} > 0$. The problem, in this case, is the computational complexity.
While Algorithm\,\ref{alg:projection} updates $\tilde{\bm{f}}$ and $\rho$ with $\bigO(\log N)$ complexity per request, the computation of  all the components of $\bm{f}$ has necessarily $\Omega(N)$ complexity. The batched operation lead to  $\bigO(N/B)$  amortized complexity.

\section{Experiments}
\label{sec:experiments}

The low complexity of our caching policy \ognew makes it a practical candidate for real-world applications. Furthermore, it allows us to test it on traces with millions of requests and millions of items, a scale that other no-regret policies have not been tested on, likely due to the impracticality of doing so in a reasonable time.
We consider a set of \emph{representative} real-world cases. \ognew's theoretical no-regret guarantees (Theorem~\ref{thm:fractional_regret}) should translate into increased robustness to variations in request patterns.

Our main reference for comparison is the optimal static allocation in hindsight (OPT), considered in the regret definition \eqref{eqn:regret}. The only no-regret policy we can evaluate on the traces we consider is the FTPL policy with noise added only at the beginning, which also enjoys $\bigO(\log N)$ complexity (see Sec.~\ref{sub:motivation}). As anticipated, FTPL suffers from high sensitivity to its configuration parameter and has difficulty tracking changes in request patterns. Finally, we also show results for LRU to demonstrate that the \ognew achieves hit ratios comparable to commonly used caching policies.

Unless otherwise stated, \ognew and FTPL are configured as required for their theoretical guarantees to hold. In particular, \ognew's learning rate $\eta$ is set according to Theorem~\ref{thm:fractional_regret}.

\subsection{Traces}
We consider a set of publicly-available block I/O traces from SNIA IOTTA repository \cite{sniaRepository}, a Content Delivery Network (CDN) trace from \cite{song2020learning, cdnRepository}, and a Twitter production cache traces from  \cite{yang2020large, twitterRepository}---see Table\,\ref{tab:trace_comparison}, the last 4 lines. From the SNIA IOTTA repository, we have considered the most recent traces---labeled as \texttt{systor}  \cite{Lee2017Understanding}---along with older traces collected at Microsoft \cite{kavalanekar2008characterization}, labeled as \texttt{ms-ex}. The \texttt{cdn} traces refer to a CDN cache serving photos and other media content for Wikipedia pages (21 days, collected in 2019). The \texttt{twitter} traces consider their in-memory cache clusters and have been collected in 2020.

Note that the full traces sometimes contain different subtraces, and each subtrace may include hundreds of millions of requests over several days. In our experiments, we consider mostly the subtraces related to Web content or e-mail servers---the only exception is the \texttt{systor} trace, which considers storage traffic from a Virtual Desktop Infrastructure. For experimental reproducibility, in Appendix\,\ref{app:traces} we report the details of the subtrace we use. 

\subsection{Results}
First, we consider the case where the cache is updated after each request ($B=1$), leaving the discussion of batched operations ($B>1$) for the next section. The main metric of interest is the hit ratio, i.e., the ratio between the number of hits and the number of requests. Typically, given a trace, the computation considers the \emph{cumulative} number of hits and requests, which is also what we presented in the results in Sec.\,\ref{sec:background}. While the traffic pattern may change over time, the cumulative representation may smooth out such variability. For this reason, in this section, we present the hit ratio computed over non-overlapping windows of $10^5$ requests, i.e., each point in the graph represents the number of hits for the last $10^5$ requests, divided by $10^5$. Unless otherwise stated, the cache size is set to 5\% of the catalog size.

\begin{figure}[th]
    \centering
    \includegraphics[width=.33\linewidth]{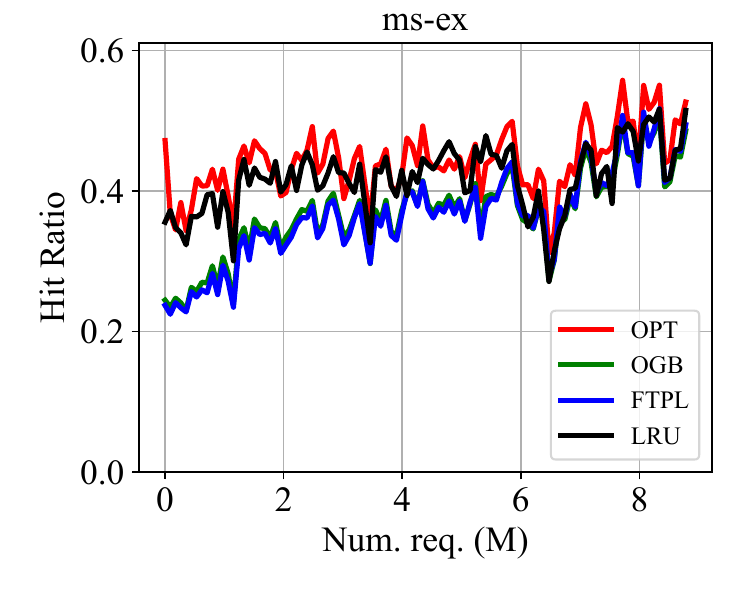}
    \includegraphics[width=.33\linewidth]{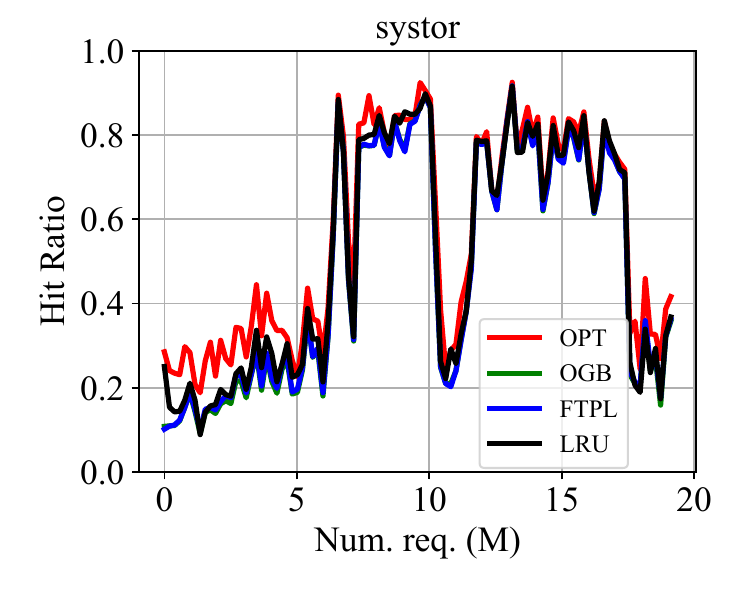}
    \caption{Real-world long (less recent) traces: Performance of \ognew policy.}
    \label{fig:old_traces_res}
\end{figure}

Figure\,\ref{fig:old_traces_res} shows the results for the less recent traces, \texttt{ms-ex} and \texttt{systor}. In both cases, we can observe a highly variable hit ratio over time for the OPT policy. LRU, FTPL and \ognew are able to follow such variability. In the \texttt{ms-ex}  trace, the FTPL and \ognew policies take some time to reach the hit ratio of OPT. This behaviour is qualitatively consistent with the fact that their time-average regret progressively vanishes (or even become negative). In other cases, such as the \texttt{systor} traces, this convergence is faster.

\begin{figure}[th]
    \centering
    \includegraphics[width=.33\linewidth]{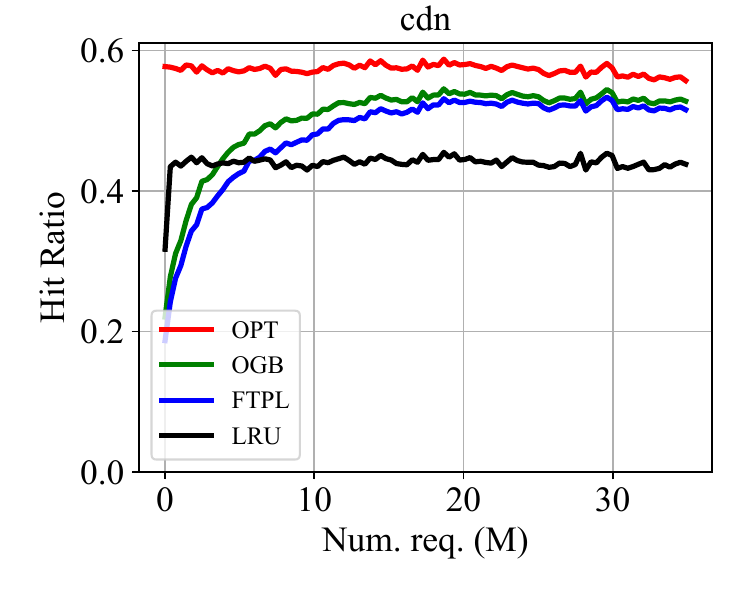}
    \includegraphics[width=.33\linewidth]{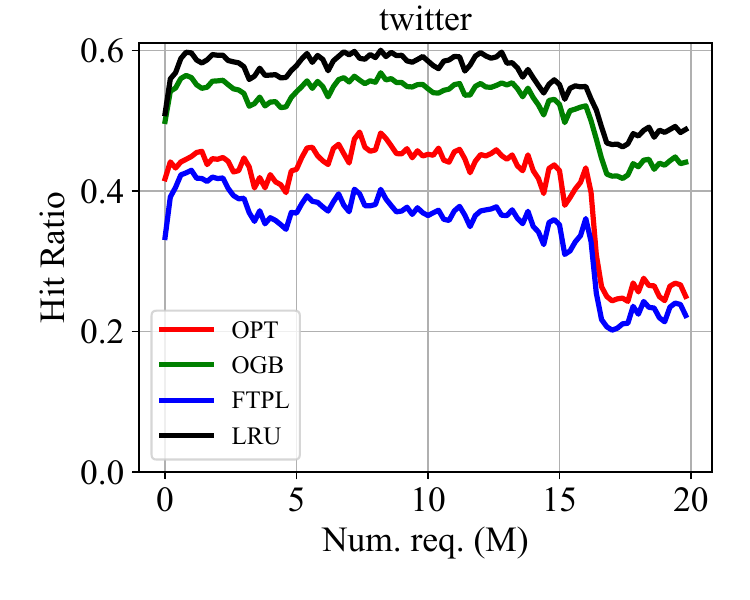}
    \caption{Real-world long (more recent) traces: Performance of \ognew policy.}
    \label{fig:recent_traces_res}
\end{figure}

A similar behavior appears in the \texttt{cdn} trace (Fig.,\ref{fig:recent_traces_res}, left). Here, the traffic pattern is much more stable, and OPT significantly outperforms the recency-based LRU, with the two no-regret policies approaching OPT. In contrast, the \texttt{twitter} trace exhibits a pattern with a higher level of temporal locality, as revealed by LRU achieving the highest hit ratio. \ognew also outperforms OPT, and its consistent performance across diverse traces like \texttt{cdn} and \texttt{twitter} demonstrates its robustness. Interestingly, FTPL approaches but does not exceed OPT's performance. FTPL is essentially a noisy LFU and is thus more suited for stationary traces.\footnote{
We tested FTPL with different values of its parameter $\zeta$ and report its best performance in Fig.,\ref{fig:recent_traces_res}.
}

\begin{figure}[th]
    \centering
    \includegraphics[width=.33\linewidth]{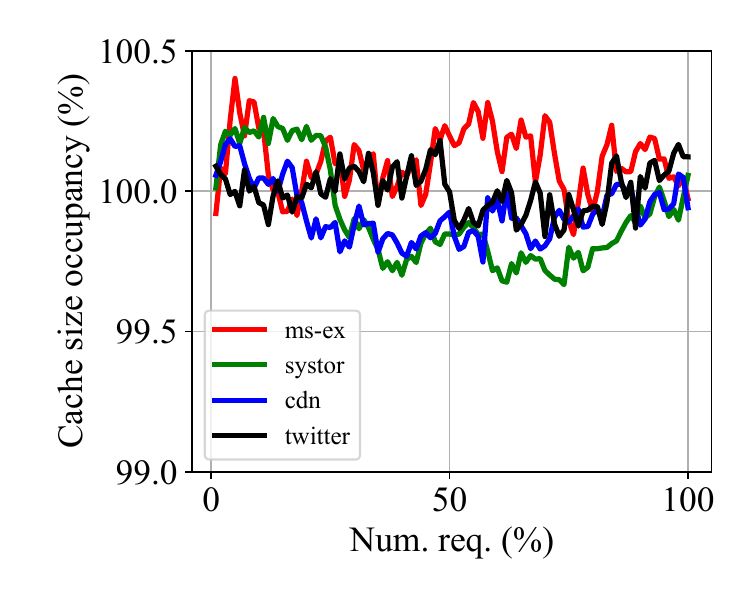}
    \includegraphics[width=.33\linewidth]{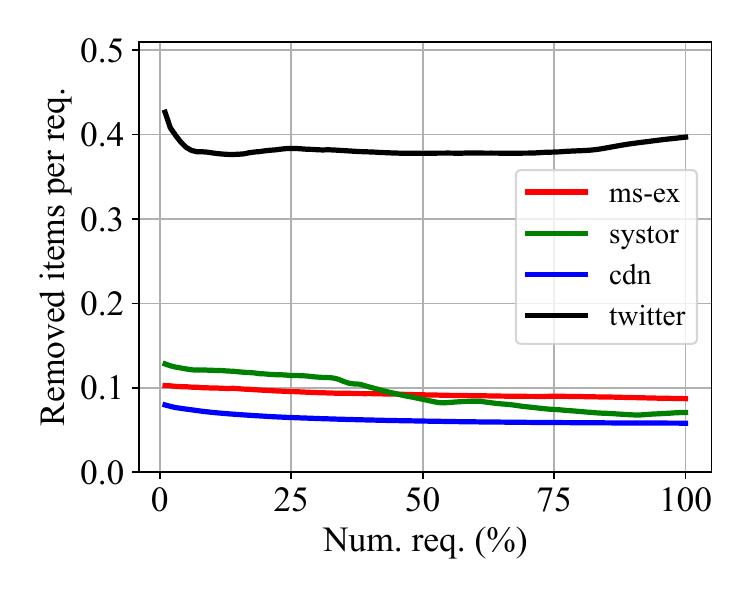}
    \caption{Cache occupancy over time (left) and average removed items per request (right).}
    \label{fig:occupancy}
\end{figure}

\vspace{1mm}
\noindent
{\bf Other statistics.} The sampling scheme we adopt does not guarantee to have exactly $C$ items in the cache. To understand the variability in the number of stored items, we record at regular intervals the cache occupancy. Figure\,\ref{fig:occupancy}, left, shows the percentage of items stored in the cache with respect to the nominal cache size. Since we have traces with different lengths, we normalized the x-axis with respect to the trace length. In all cases, the variability is limited to 0.5\% of the cache size. Therefore, it is still possible to use \ognew under hard cache constraints as far as we set the expected cache size to be slightly inferior to the actual storage available.

Figure\,\ref{fig:occupancy}, right, shows the average number of items that are removed from $\tilde{\bm{f}}$ in Algorithm\,\ref{alg:projection} (lines \ref{alg:repeat}--\ref{alg:repeat_end}) over  windows of $10^5$ requests.
In all cases we are below 0.5 removed items per requests, confirming the theoretical analysis  at the end of Sec.\,\ref{sec:projection}.

\subsection{Batched arrivals and fractional case}
\label{sub:batched}
\begin{figure}[th]
    \centering
    \includegraphics[width=.33\linewidth]{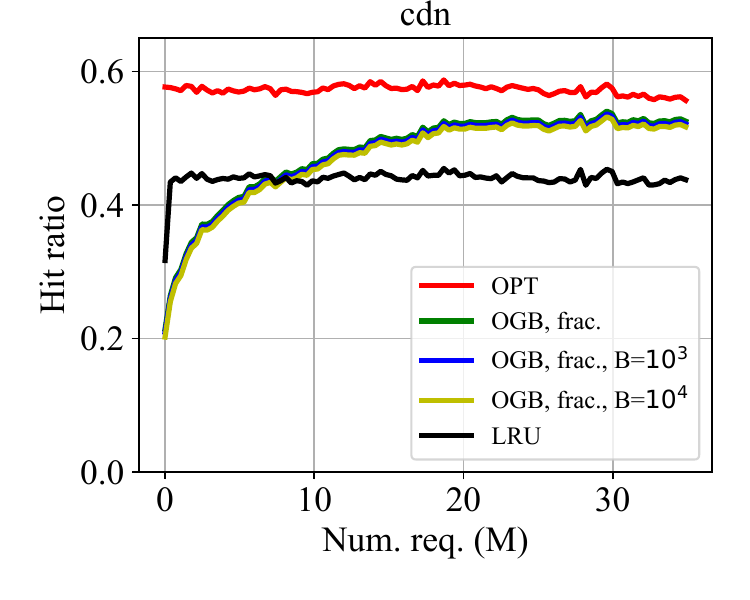}
    \includegraphics[width=.33\linewidth]{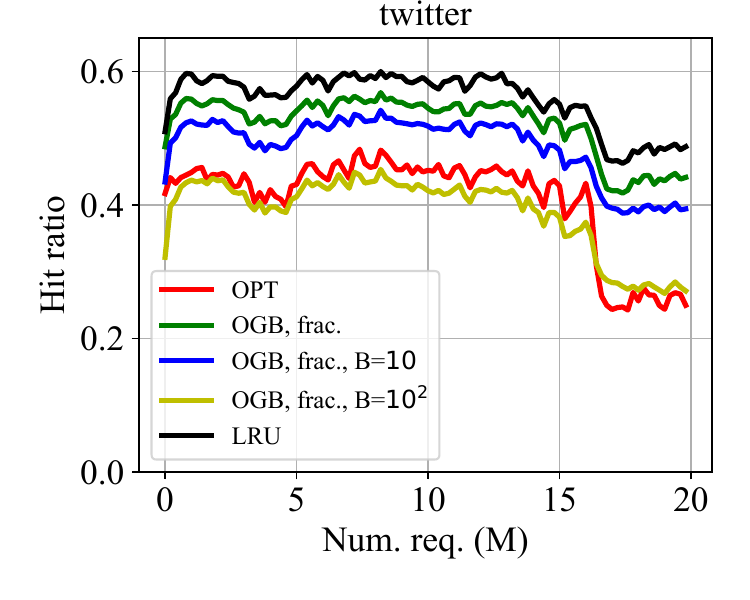}
    \caption{Real-world long traces: Performance of \ognew policy in the fractional case for variuos values of the batch size $B$.}
    \label{fig:batched_res}
\end{figure}

We consider now a batched operation and evaluate the impact of $B$, the number of requests in the batch, on the hit ratio for the two most recent traces, \texttt{cdn} and \texttt{twitter}. We perform our experiments both in the integral and in the fractional setting. The hit ratios in the two cases are practically indistinguishable. This confirms the intuition that, for large catalogs and cache sizes,  storing fractions of the items or using the fractions as probabilities to sample the items. We report then only the results for the fractional setting in Fig.\,\ref{fig:batched_res}.

For the \texttt{cdn} trace, large values of $B$ do not affect the performance, while for the \texttt{twitter} trace even small batches of 100 requests have a significant impact on the hit ratio. The main reason lies in the temporal locality of the requests, \ie, how concentrated are the requests for each item. If the requests are highly concentrated, \eg, a burst of few requests for the same item in less than $B$ consecutive requests, then batching the requests together affects negatively  the hit ratio. The cache, in fact, is updated after the batch, but that item is not requested in the following batches. Clearly for popular items that are continuously requested this effect does not appear. But the presence of many less popular items requested in burst has a significant impact.

We have analyzed in detail the \texttt{twitter} trace in  Appendix\,\ref{app:locality} and observed that there are a set of items that are requested in short bursts but still account for up to 20\% of the hit ratio. In the \texttt{cdn} trace, items with similar request patterns have much less influence, which explains why the results are less affected by $B$.

Overall, while the parameter $B$ helps decrease computational complexity in the fractional case, its optimal configuration depends on the trace characteristics. It should be adjusted (even dynamically) to account for the impact of short request bursts.

\section{Related work}
\label{sec:related}

The caching problem has been extensively studied in the literature. Since our work concern the efficient implementation of a policy with specific characteristics, we discuss here the literature considering these aspects.

Well known caching policies, such as LRU, LFU \cite{matani20211}, FIFO \cite{eytan2020s} and ARC \cite{megiddo2003arc} are widely adopted due to their constant complexity: at each new requests they update the cache in $\bigO(1)$ steps. In some contexts where the arrival rate is extremely high and the speed of the cache is a key issue, these policies are the only viable solution. The price paid is represented by their performance in terms of hit ratios. These policies work well with specific traffic patterns, but do not provide regret guarantees.

Another class of policies, which includes GDS \cite{cao1997cost}, trade  computational efficiency for often higher hit ratios (GDS has $\log C$ time complexity). Although they are able to adapt to different contexts, they do not provide regret guarantees. This is true also for policies that are based on Machine Learning tools to predict future request, and therefore improve the performance \cite{rodriguez2021learning}. Such predictions are based on past observations, and no guarantees are provided. 

Inspired by the online optimization framework, policies that are able to provide theoretical guarantees on their performance without any assumption on the arrival traffic pattern have been proposed recently. Most of the works consider the fractional case: in this scenario, as discussed in Sec.\,\ref{sec:fractional}, there is an intrinsic linear complexity, so all the solutions have at least $\bigO(N)$ complexity. This is the case for example of the policies in \cite{paschos2020online}---which showed that \og in \cite{paschos2019learning} may be implemented with $\bigO(N)$ complexity---as well as those in \cite{si2023no}. In the integral setting, it is common to sample the cache content starting from a fractional solution computed by a gradient method. A naive implementation leads then to $\bigO(N)$ complexity, as discussed in Sec.\,\ref{sec:main_idea}. Another possibility is to use a variant of the FTL approach~ \cite{mukhopadhyay2021online}. While the original proposal has $\bigO(N \log N)$ complexity \cite{bhattacharjee2020fundamental}, these variant that associates a single initial noise \cite{mukhopadhyay2021online, mhaisen2022optimistic, faizal2022regret} have $\bigO(\log N)$ complexity. These works, despite using a low complexity scheme, provide experimental results for traces with at most $10^4$ items and $10^5$ requests, which may not capture the complexity of  real-word request patterns. Our experiments with longer traces and larger catalogs highlight some limits of FTPL.

Differently from all the above works, our solution is computationally efficient, with  $\bigO(\log N)$ complexity, and, since it is based on a robust online gradient approach,  can be used in real-world scenarios, making it a practical caching policy. 

Our solution combines low complexity ($\bigO(\log N)$) and robustness to different request patterns, making it a practical caching policy. 

\section{Conclusions and perspectives}
\label{sec:conclusion}

No-regret caching policies are able to offer performance guarantees with no assumption on the arrival traffic pattern. Their computational complexity has limited their application to small scale, unrealistic experiments, where the requested items belongs to a limited catalog.

In this paper we provide aa new variant of the gradient-based online caching policy. We coupled the maintenance of the caching probabilities for each item with a sampling scheme that provides coordinated samples, thus minimizing the replacement in the cache. 
Thanks to its low complexity, we were able to test the online gradient based caching policy in real-world cases, showing that it can scale to long traces and large catalogs. 

As future work, we will explore the possibility of extending our policy to cases where items have different sizes and introducing correlation in the items' sampling scheme to further reduce the variability of the cache's instantaneous occupancy. Additionally, we will investigate the applicability of our solution to more general online convex optimization problems.

\newpage
\bibliographystyle{ACM-Reference-Format}
\bibliography{00_refs}

\newpage

\appendix

\section{Proof of Theorem~\ref{thm:fractional_regret}}
\label{sec:proof_regret}

\begin{proof}
We prove the result for a general online convex optimization problem with $L$-Lipschitz convex cost functions $c_t: \mathcal F \to \mathbb R$ over the bounded convex set $\mathcal F \subset \mathbb R^N$ with $\radius(\mathcal F)= \min_{\f \in \mathcal F} \sup_{\hat{\f}\in \mathcal F} \lVert \f - \hat{\f} \rVert$.\footnote{
    We observe that here we refer to costs rather than rewards. 
} 

Let $\mathcal A$ denote the standard online gradient descent algorithm~\cite{zinkevich03}: it updates the state at each time slot as follows (this equation corresponds to \eqref{eqn:ogbnew_update}):
\begin{equation*}
\bm{f}_{t+1} = \Pi_\mathcal{F} \left( \bm{f}_{t} - \eta \nabla  c_t(\bm{f}_t) \right),
\end{equation*}
and experiences a cost $c_{t+1}(\bm{f}_{t+1})$.

The algorithm $\mathcal A$ enjoys the following (fracational) regret guarantees for $\f_0 = \arg\min_{\f\in \mathcal F} \sup_{\hat{\f}\in \mathcal F} \lVert \f - \hat{\f} \rVert$ and $\eta = \frac{\radius(\mathcal F)}{L \sqrt{T}}$:
\begin{align}
    R_T(\mathcal A) & \triangleq \sup_{c_0, c_1, \dots c_{T-1}}\left\{\sum_{t=0}^{T-1} c_t(\f_t) - \min_{\f \in \mathcal F} \sum_{t=0}^{T-1} c_t(\f)\right\}\\
    & \le \frac{\radius(\mathcal F)^2}{2 \eta} + \frac{\eta L^2 T}{2}\nonumber\\
    & = \radius(\mathcal F) L \sqrt{T}. 
\end{align}
Now, we consider an algorithm $\mathcal A'$ which selects the same state as $\mathcal A$ at time slots multiple of $B$ and freezes the state at other time slots, i.e., $\f'_t = \f_{\ell(t)}$, where $\ell(t)\triangleq\lfloor t/B \rfloor B$. 

We compute the difference in the total cost experienced by the algorithms $\mathcal A$ and $\mathcal A'$.

\begin{align}
    c_t(\f'_t) - c_t(\f_t) & = c_t(\f_{\ell(t)}) - c_t(\f_t)\\
    & \le L \lVert \f_t - \f_{\ell(t)} \rVert\\
    & \le L \sum_{\tau = 0}^{t - \ell(t)-1} \lVert \f_{\ell(t)+\tau+1} - \f_{\ell(t)+\tau} \rVert\\
    & \le L \sum_{\tau = 0}^{t - \ell(t)-1} \lVert  \Pi_\mathcal{F}\left(\f_{\ell(t)+\tau} - \eta \nabla c_{\ell(t)+\tau}\right) - \f_{\ell(t)+\tau} \rVert\\
    & \le L \sum_{\tau = 0}^{t - \ell(t)-1} \lVert  \f_{\ell(t)+\tau} - \eta \nabla c_{\ell(t)+\tau} - \f_{\ell(t)+\tau} \rVert\\
    & \le \eta L^2 (t- \ell(t)).
\end{align}
Summing over $T$ we obtain
\begin{align}
    \sum_{t=0}^{T-1} c_t(\f'_t) - c_t(\f_t) & \le \eta L^2 \left(\left\lfloor \frac{T-2}{B} \right\rfloor \frac{B(B-1)}{2} + \sum_{t=\ell(T-1)}^{T-1} (t- \ell(t))\right)\\
    & = \eta L^2 \left(\left\lfloor \frac{T-2}{B} \right\rfloor \frac{B(B-1)}{2} + \frac{(T-1- \ell(T-1))(T- \ell(T-1))}{2}\right)\\
    & \le \eta L^2 \left(\left\lfloor \frac{T-2}{B} \right\rfloor \frac{B(B-1)}{2} + \frac{(B-1)(T- \ell(T-1))}{2}\right)\\
    & \le \eta L^2 \frac{B-1}{2} \left(\left\lfloor \frac{T-1}{B} \right\rfloor B + T- \ell(T-1)\right)\\
    & \le \eta L^2 T \frac{B-1}{2}
\end{align}
Then algorithm $\mathcal A'$ enjoys the following regret bound for $\f_0 = \arg\min_{\f\in \mathcal F} \sup_{\hat{\f}\in \mathcal F} \lVert \f - \hat{\f} \rVert$ and $\eta = \frac{\radius(\mathcal F)}{L \sqrt{T B}}$:
\begin{align}
    R_T(\mathcal A') & \le \frac{\radius(\mathcal F)^2}{2 \eta} + \frac{\eta L^2 T}{2} + \eta L^2 T \frac{B-1}{2}\\
    & = \frac{\radius(\mathcal F)^2}{2 \eta} + \frac{\eta L^2 T B}{2}\\
    & = \radius(\mathcal F) L \sqrt{T B}. 
\end{align}

The result for the caching problem is obtained by considering that $\radius(\mathcal F) = \sqrt{C\left(1- \frac{C}{N}\right)}$ and $L=1$.

\end{proof}

\section{Additional information on the results}

\subsection{Details on the used traces}
\label{app:traces}

For experimental reproducibility, we report here the details of the subtrace we use. The \texttt{ms-ex} is a trace taken from \cite{kavalanekar2008characterization} named ``Microsoft Enterprise Traces, Exchange Server Traces,'' which has been collected for Exchange server for a duration of 24-hours---we consider the first 3.5 hours. The \texttt{systor} traces \cite{Lee2017Understanding} collect requests for different block storage devices over 28 days: we consider 12 hours (March, 9th) for the device called ``LUN2.'' The \texttt{cdn} trace contains multiple days of traffic, of which we consider portions of 6 hours---we tested different intervals finding similar qualitative results. Finally, for \texttt{twitter} we considered the first 20 millions requests from cluster 45. Also in this case, we obtained similar qualitative results for other clusters.

\subsection{Analysis of the request temporal locality}
\label{app:locality}

In order to understand the impact of the batch size on the performance, we need to analyze the characteristics of the arrival pattern. In particular, we consider the \emph{lifetime} of the items, \ie, the difference between the timestamps of the last and first requests for each item. If we have an infinitely large cache, then each item contributes with a number of hits that is equal to the number of requests minus 1: this is because, in policies like LRU, the first request is always a cold miss, and it is used to bring the item in the cache, and all the following requests generate hits---the actual number of hits depends on the cache size, but with an infinite cache, the item is never evicted, so this represents an upper bound. We then sort the items by their lifetimes, and cumulatively compute the hit ratio they generate.

\begin{figure}[ht]
    \centering
    \includegraphics[width=.33\linewidth]{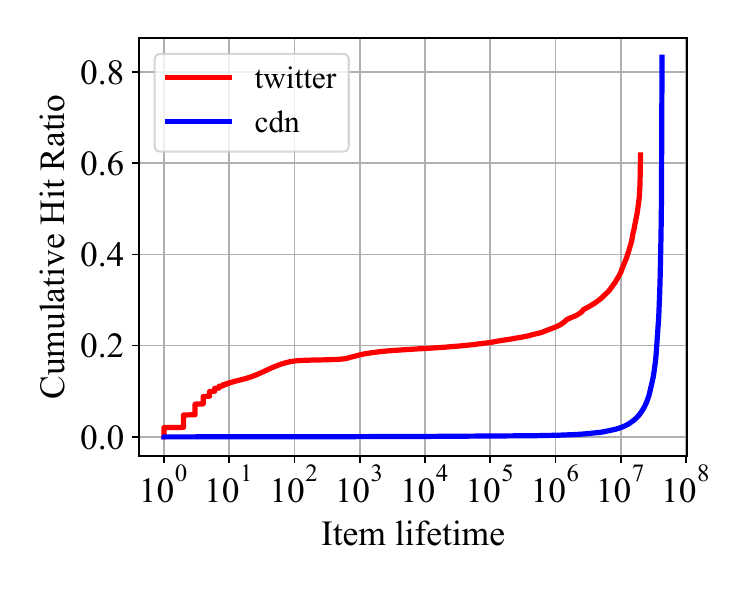}
    \includegraphics[width=.33\linewidth]{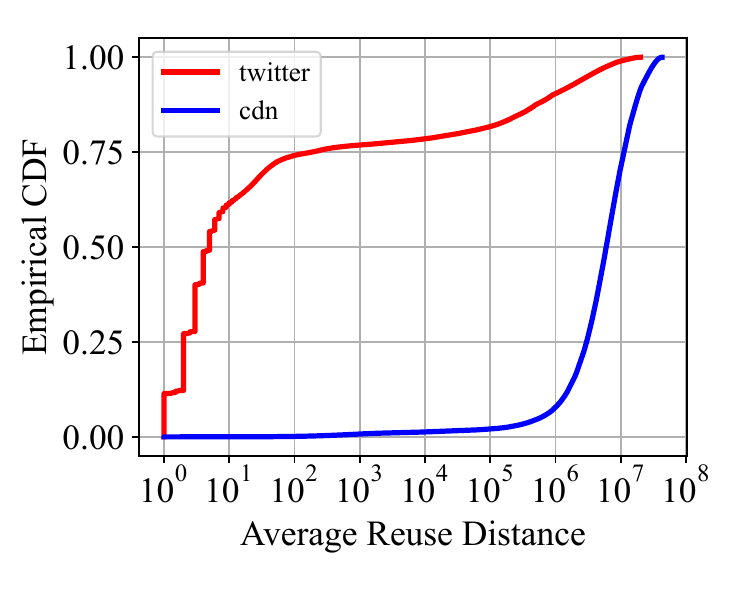}
    \caption{Cumulative (maximum) hit ratio provided by items sorted by their lifetime (left), and empirical CDF of the reuse distance (right).}
    \label{fig:locality_analysis}
\end{figure}

Figure\,\ref{fig:locality_analysis} (left) shows that the set of items with lifetime smaller than 100 requests (recall that the timestamp is actually determined by the progressive number of received requests, independently from which item is requested) for the \texttt{twitter} trace accounts for almost 20\% of the hit ratio. Therefore, if a batch size is bigger than the item lifetime, that item will not generate any hit, because its set of requests is absorbed by the batch. By comparison, in the \texttt{cdn} trace, item have a much larger lifetime, \ie, they are continuously requested. 

For both traces, we have also computed the average \emph{reuse distance}, \ie, the average timestamp difference between two consecutive requests for an item. In Fig.\,\ref{fig:locality_analysis}, right, we show the empirical Cumulative Distribution Function (CDF) of the reuse distance. In the \texttt{cdn} trace, the reuse distance for most of the items is large: this, combined with the large lifetime, indicates that the items are regularly requested throughout the whole trace, which represents the ideal scenario for a scheme that collects batches of requests. On the contrary, in the \texttt{twitter} trace a large percentage of items requested with a small reuse distance, which favors recency-based caching schemes, but limits the  benefit of grouping the requests in batches. 

\end{document}